\documentclass{vgtc}                          

\ifpdf
  \pdfoutput=1\relax                   
  \usepackage{graphicx}                
  \DeclareGraphicsExtensions{.pdf,.png,.jpg,.jpeg} 
\else
  \usepackage{graphicx}                
  \DeclareGraphicsExtensions{.eps}     
\fi%

\graphicspath{{figures/}{pictures/}{images/}{./}} 

\usepackage{microtype}                 
\PassOptionsToPackage{warn}{textcomp}  
\usepackage{textcomp}                  
\usepackage{mathptmx}                  
\usepackage{times}                     
\usepackage{cite}                      
\usepackage{tabu}                      
\usepackage{booktabs}                  
\usepackage[ampersand]{easylist}
\usepackage{color}
\usepackage[normalem]{ulem}
\usepackage {gensymb} 

\onlineid{0}

\vgtccategory{Research}

\vgtcinsertpkg

\title{CloudFindr: A Deep Learning Cloud Artifact Masker for Satellite DEM Data}

\author{Kalina Borkiewicz\thanks{e-mail: kalina@illinois.edu} %
\and Viraj Shah\thanks{e-mail: vjshah3@illinois.edu} %
\and J.P. Naiman\thanks{e-mail: jnaiman@illinois.edu} %
\and Chuanyue Shen\thanks{e-mail: cs11@illinois.edu} %
\and Stuart Levy\thanks{e-mail: salevy@illinois.edu} %
\and Jeff Carpenter\thanks{e-mail: jdcarpen@illinois.edu} %
} %
\affiliation{\scriptsize University of Illinois at Urbana-Champaign}

\abstract{
Artifact removal is an integral component of cinematic scientific visualization, and is especially challenging with big datasets in which artifacts are difficult to define. In this paper, we describe a method for creating cloud artifact masks which can be used to remove artifacts from satellite imagery using a combination of traditional image processing together with deep learning based on U-Net. Compared to previous methods, our approach does not require multi-channel spectral imagery but performs successfully on single-channel Digital Elevation Models (DEMs). DEMs are a representation of the topography of the Earth and have a variety applications including planetary science, geology, flood modeling, and city planning.

} 

\begin{document}


\firstsection{Introduction}

\maketitle

Cloud detection in satellite imagery is a problem that has plagued scientists for decades (e.g.\cite{clouddetection1988, noaa1993, modis2004, fmask2012, mcnet2021}). Whether a scientist's area of research is the clouds themselves or the land beneath them, it is useful to separate the two classes of objects, though a universal method for doing so remains elusive. Various methods have been proposed depending on the type of data (e.g. spectral \cite{JEPPESEN2019247, wu2016cloudfinder}, time-evolving   \cite{timeevolving2012, timeevolving2013}) and the specific research objective.

However, as of this writing no current cloud detection methods exist for when the objective is not scientific data analysis, but rather cinematic scientific visualization which aims to create aesthetically pleasing imagery for a general audience. A visualization created for outreach purposes requires a different focus in the underlying data processing in order to create a result that is not only understandable, but also visually appealing to the general public. Aesthetically-pleasing visualizations are both more educational \cite{cawthon2007} and are perceived to be more credible than those which are not \cite{misinformation5}. 

Our work differs from other cloud detection methods primarily in two ways -- the underlying data, which is limited to a 1-dimensional elevation model rather than 3+-dimensional spectral imagery; and the general purpose. The aim of our work is not data cleaning for purposes of data analysis, but rather to create a cinematic scientific visualization which enables effective science communication to broad audiences. Great care must be applied in visualizations of complex data for lay audiences, and additional data processing, camera choreography, and different methods of rendering are required to achieve a goal of clear communication \cite{chromatophore}.

The CloudFindr method described here can be used to algorithmically mask the majority of cloud artifacts in satellite-collected DEM data by visualizers who want to create content for documentaries, museums, or other broad-reaching science communication mediums, or by animators and visual effects specialists who want to use such DEM data to create realistic landscapes and backdrops in otherwise fictional computer-generated movie scenes.

\begin{easylist}[itemize]
\end{easylist}

\subsection{Cinematic Scientific Visualization in \textit{Atlas of a Changing Earth}}

\begin{figure}[tb]
 \centering
 \includegraphics[width=\columnwidth]{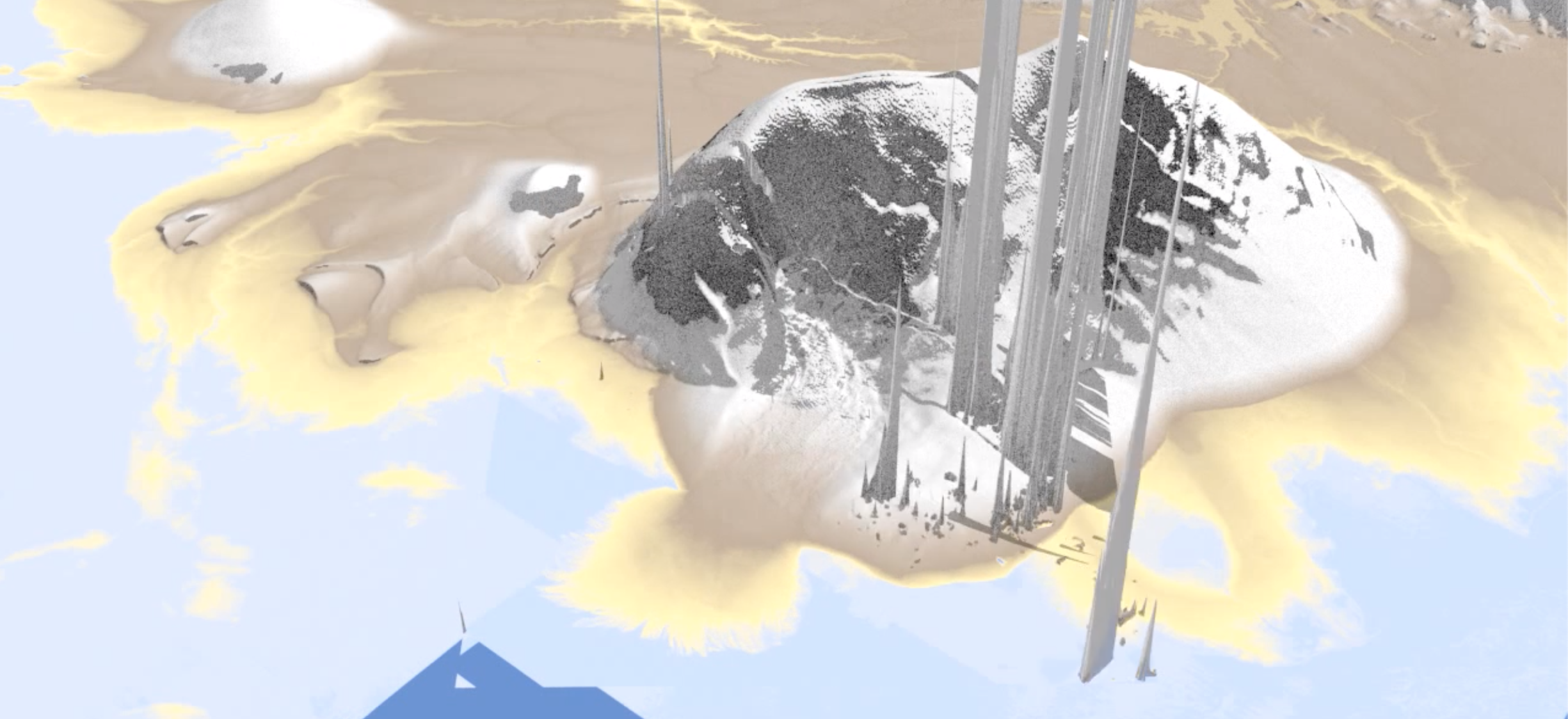}
 \caption{A 3D visualization of a DEM region without cloud artifact removal, showing large spikes where the height of the land is incorrectly labelled with the height of a cloud.}
 \label{fig:spikes}
\end{figure}

When creating a public-facing outreach visualization for broad public distribution via films shown in giant immersive theaters (e.g. planetarium domes, IMAX screens), it is critical that data must be artifact-free. If the dataset in question is a digital elevation model (DEM) of land, clouds are considered to be artifacts and must be removed. A single cloudy DEM pixel, reprojected into 3D, would result in an unacceptable massive spike in the landscape that is sure to draw audience attention away from the immersive experience of the story (see Figure \ref{fig:spikes}), especially on a $75$+ foot screen.

The Advanced Visualization Lab (AVL) at the National Center for Supercomputing Applications encountered this problem when working on a documentary, \textit{Atlas of a Changing Earth}, which features three locations visualized from the ArcticDEM dataset\cite{arcticdem_data}. The motivation for the work described in this paper was the time-consuming manual cloud removal that was required in order to create a seamless, smooth, artifact-free cinematic visualization of the DEM data. Though some basic automatic cloud-removal techniques were used during the making of the documentary, they were not satisfactory, and the process still required weeks of manual review.

\begin{figure}[tb]
 \centering
 \includegraphics[width=\columnwidth]{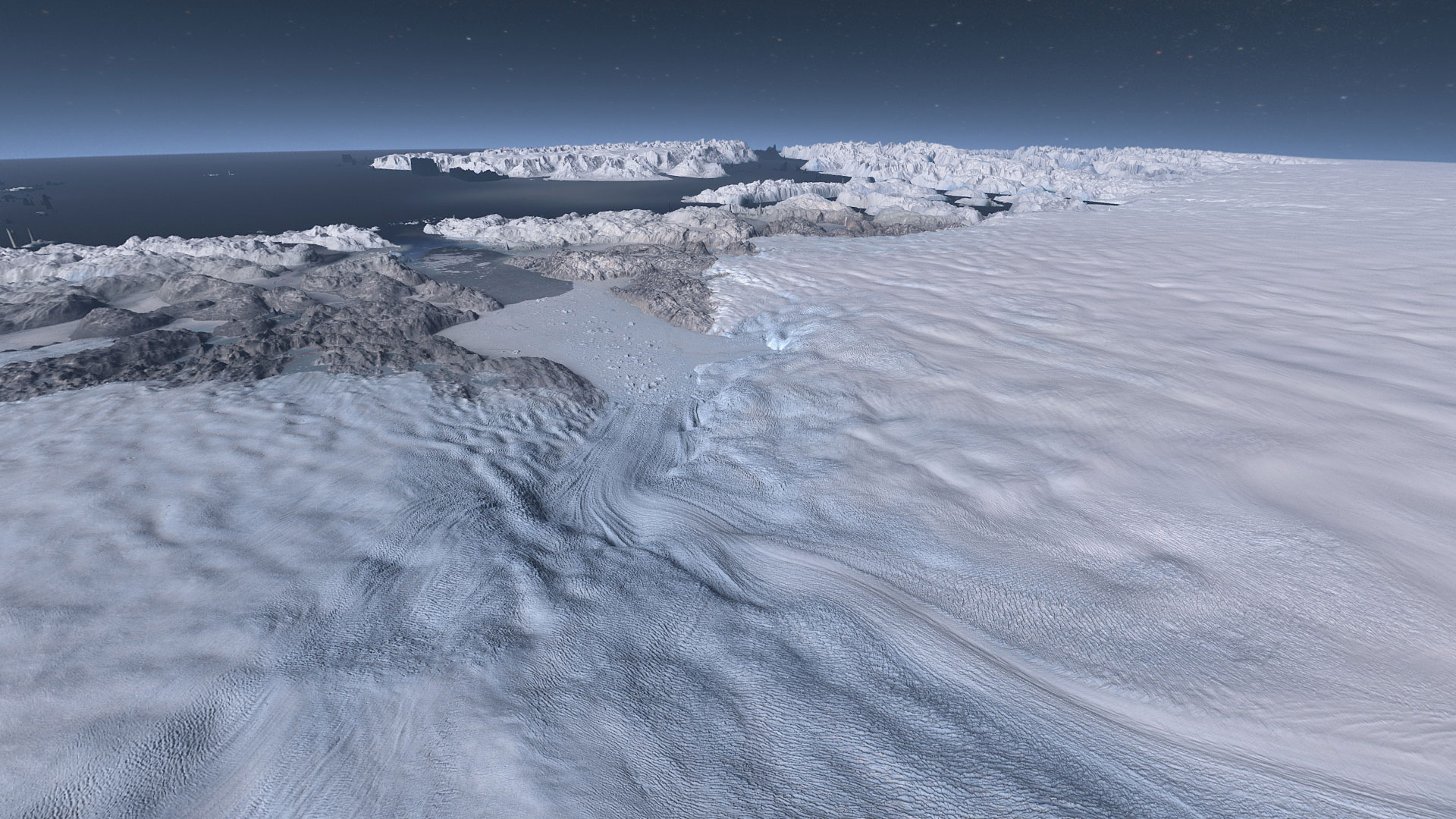}
 \caption{Final cloud-free cinematic rendering of the Jakobshavn glacier used in the \textit{Atlas of a Changing Earth} documentary.}
 \label{fig:final}
\end{figure}

\subsection{Cloudy DEMs}

Digital elevation model data is represented as imagery, where each pixel holds a value for the elevation at that location. DEM data is often gathered by satellite, and in these cases, clouds become an issue. The annual global mean cloud cover is approximately 66\% \cite{cloudcover}, so Earth-facing satellites which aim to study the land collect noisy or incorrect data when clouds obstruct their view. Making this data more usable requires significant data cleaning. For many scientific purposes, it is useful to retain as much of the collected data as possible during data cleaning, even if there is a small number of artifacts. In other words, there is higher value in the precision of the data than in the recall. For purposes of cinematic visualization, the opposite is true - recall is more important than precision. It is preferable to lose some correct data in the process of data cleaning to ensure the removal of the artifacts, which are not only visually unappealing but also inaccurate representations of the data.

The ArcticDEM dataset covers all land north of 60\degree, and it redefined the arctic from the worst to one of the best-mapped regions of the Earth \cite{ArcticDEM_EGU}. The data is collected by the WorldView $1$-$3$ satellites and GeoEye-$1$ satellite, and is processed to remove clouds and other errors, with an absolute error of $<$0.5 meters \cite{arcticdem_data}. The ArcticDEM project makes their derived DEMs readily available, but intellectual property considerations prevent release of the original optical imagery, hence the need for the innovations in this paper. The ArcticDEM data is collected in ``strips'', with each strip being a snapshot of a particular area. This can be thought of as a puzzle piece. Over time, the various puzzle pieces are gathered, and eventually there are enough to put together a complete puzzle. Because the strips are gathered at different points in time, putting them together does not create one single, seamless, final mosaic, but rather, puzzle pieces are periodically updated and replaced. This is where the puzzle analogy starts to break apart - a strip may cover some of the same area as a previous strip, but does not exactly ``replace'' a previous puzzle piece, as it may not have the same shape and coverage.

To build a cloud-free, complete mosaic for the \textit{Atlas of a Changing Earth} documentary, cloud masks were manually created for each strip (described in Section \ref{section:gt_creation}) and multiplied against the data to remove the artifacts. The cloud-free strips were then accumulated to build up the mosaic, at which point the visualization video begins. The strips continue to update throughout the visualization. 

Cloud detection was a manual, time-consuming process during the documentary production, however, it produced a valuable output in addition to the visualization itself - a large collection of labelled data. Detecting clouds in DEM data has a unique set of challenges: clouds may be be a small cluster of pixels or may cover the whole strip and beyond; strips that have hard edges may cut through features, so there is no guarantee that even a cumulus cloud is complete and has an organic, recognizable outline; haze and clouds that are low to the ground may be difficult to distinguish from land and may create noise which is not otherwise identifiable as ``clouds''; and there is only a single channel of data per pixel, unlike in multispectral imagery, which is most commonly used for cloud detection \cite{cloud-survey}.

\section{Related Work}

Cloud detection is a specific application of the broader field of anomaly detection with methods spanning different techniques and applications. Techniques range from information theoretic to classification-based to statistical; applications span cyber-intrusion detection to image processing to sensor networks\cite{anomaly-survey}. Deep learning methods can be applied to anomaly detection using algorithms that are supervised, unsupervised, hybrid, or one-class neural networks \cite{anomaly-survey-deep}. An issue when attempting anomaly detection with spatiotemporal data is that there is often a lack of a clear boundary between normal and abnormal cases \cite{cao2018} -- in the case of cloud detection, it can be difficult to determine if a pixel contains a cloud, or a snow-peaked mountain.

Much research on cloud detection in particular focuses on spectral imagery as input data, rather than DEM input. Cloud detection methods for these data are based on cloud optical properties and may detect cloud/no-cloud, cloud/snow, and/or thin/thick cloud regions of an image \cite{cloud-survey}. Fmask \cite{fmask2012} is a popular algorithm for detecting both clouds and cloud shadows in spectral imagery. A recent paper by Wu, et al \cite{wu2016cloudfinder} uses DEM data, but for validation of their spectral cloud-finding results, rather than for the detection directly.

The method described in this paper uses deep learning image segmentation to detect and mask out cloud regions. This is based on the popular U-Net algorithm \cite{unet}, initially developed for medical image segmentation but which has since been adopted for use in other fields that require classifying image pixels. The RS-Net \cite{JEPPESEN2019247} and MC-Net \cite{mcnet2021} methods also use U-Net for cloud detection, but once again on spectral imagery rather than DEM data. Other notable recent machine learning image segmentation papers based on U-Net include a method for identifying vortex boundaries in scientific visualizations \cite{berenjkoubvortex1} and a method for removing clouds in $3$-channel RGB spectral imagery with generative adversarial networks \cite{cloudRemovalGAN}.

\section{Method}
\begin{easylist}[itemize]
\end{easylist}

\subsection{Ground Truth Mask Creation}\label{section:gt_creation}

\begin{figure}[tb]
 \centering
 \includegraphics[width=\columnwidth]{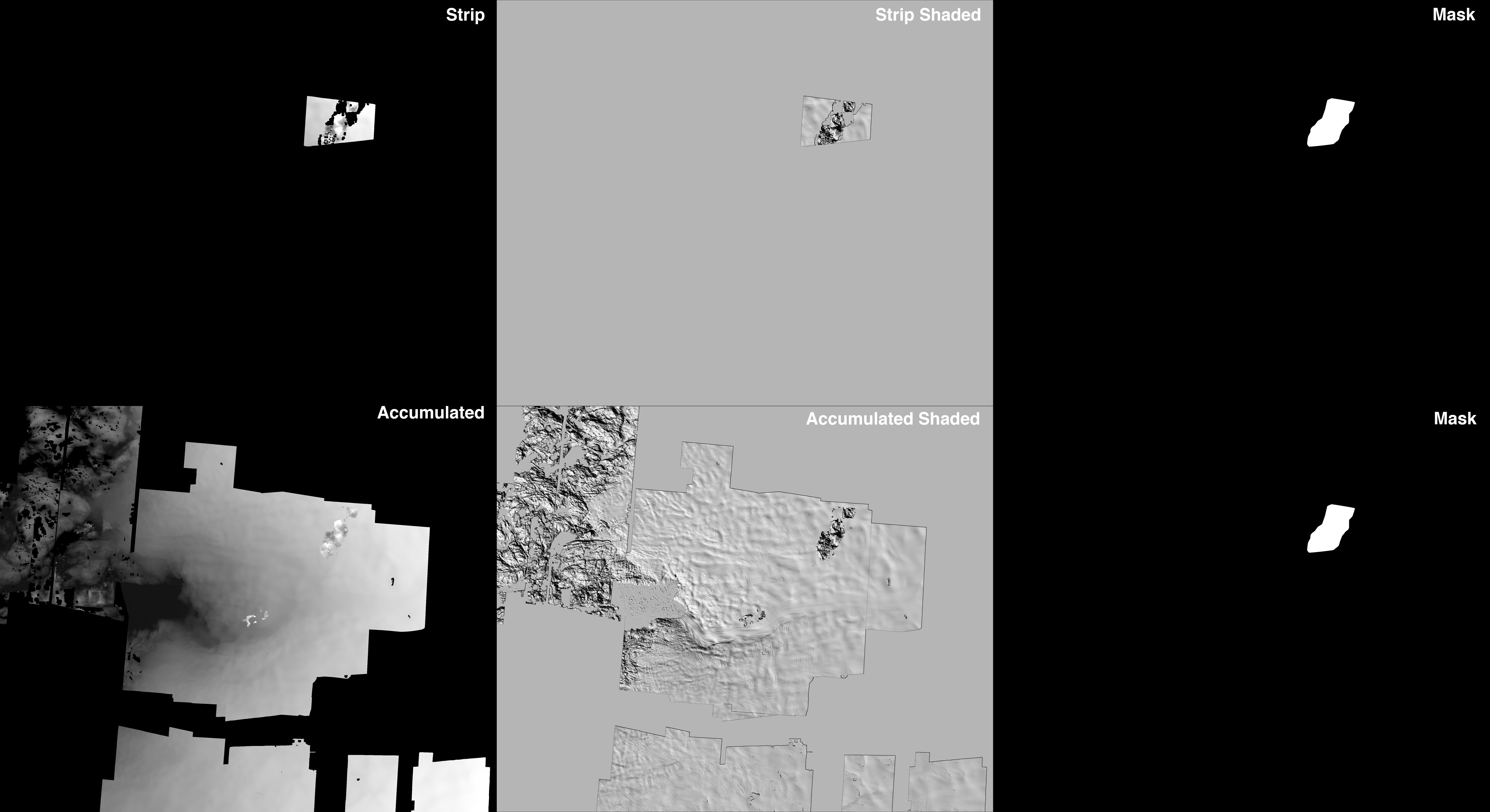}
 \caption{Example showing the inputs (left, middle) used to output a hand-drawn mask (right) for one sample timestep. Top row shows individual strips, bottom row shows accumulated buildup of strips. Left column shows DEM data, middle column shows artificially shaded preview, right column shows resulting mask (repeated in both rows).}
 \label{fig:dataTypes}
\end{figure}

The labelled dataset used as the ground truth in training was created as a byproduct of the work toward the documentary \textit{Atlas of a Changing Earth}, co-produced by Thomas Lucas Productions and the Advanced Visualization Lab at the National Center for Supercomputing Applications. The artifacts were masked and removed manually in order to fit the timeline of the film production, and these resulting masks served a secondary purpose as the inputs to our machine learning model. 

The first step in acquiring the data was identifying an area of interest and downloading a subset of the data at a suitable resolution. A $3473$x$2840$ pixel region was initially selected around the Jakobshavn glacier, a $110,000$-square km glacier in Greenland, and serves as our dataset. GEOTIFF images were downloaded from the ArcticDEM website and aligned using the georeferenced imagery, so that each new data strip would be in the correct pixel location within our selected region of interest. Several derivative versions of the data were created: ($1$) images that show one strip at a time and leave the rest of the frame blank; ($2$) images that are an accumulation of strips up until the current timestep; ($3$) images where each pixel corresponds to the time that an accumulated pixel was added; and ($4$) images that are artificially-shaded using gdaldem's ``hillshade'' mode\footnote{https://gdal.org/programs/gdaldem.html} for easier visual inspection; among others.

A multimedia specialist on the team used the software Nuke\footnote{https://www.foundry.com/products/nuke} to visually inspect the individual DEM strips, comparing them with strips gathered immediately before and after to identify and manually mask out areas that appeared to be artifact-ridden. Using a visual effects technique called rotoscoping, in which a vector mask is created in one image frame and filled in with imagery from another, the expert drew the masks for each new data strip by comparing the various images described above over time, interactively making adjustments to image intensity as needed for better visual acuity. Figure \ref{fig:dataTypes} shows a sample of types of inputs into this manual process as well as the output mask for a single timestep.

The hand-drawn masks were not pixel-precise, but were overdrawn for reasons of convenience - e.g. if $90$\% of a strip was cloud-covered, it was more time-efficient to mask out the whole strip rather than finding the individual pixels that were valid. This was satisfactory for purposes of the documentary, but would not be suitable for a machine learning task. We therefore created a second set of ``motion masks'' where each pixel contained a $1$ only if the pixel had been updated (moved) in that current timestep, and 0 otherwise, based on derivative data version ($3$) described above. Multiplying these two masks together clipped the expert-created overdrawn masks to only pixels that were present in the strip at that timestep. The resulting masks are both expert-driven and pixel-precise. 

\subsection{Data Pre-Processing}
Data must be processed prior to being used for training in order to optimize training time and results. First, each image and its corresponding ground-truth mask is subdivided into patches of size $224$x$224$ pixels. This size was chosen in order to divide cleanly into whole numbers when downsampled with the U-Net algorithm. Other patch sizes were tested during parameter tuning, ranging from roughly $100$x$200$ - $600$x$600$, and this size was chosen for having a good ratio of processing speed to manageable number of output images. Patches were set to overlap one another by $50$ pixels to account for artifacts around the borders of the image, which are known to occur with many Convolutional Neural Network-based image processing algorithms \cite{JEPPESEN2019247}. This also had the result of creating more training data with different patch croppings. The value of $50$ pixels was selected by visually inspecting a sampling of predicted output masks and determining the region of consistently-inaccurate predictions around the borders. Because clouds are more rare than non-clouds in the data and they are the subject of interest, only the patches that had at least one pixel of cloud (as determined by the ground-truth mask) were saved. There were originally $978$ images of size $3473$x$2840$, which were converted into $4399$ patches of size $224$x$224$. Scripts were developed for splitting the full-sized image into patches and for reassembling the patches into a full-size image.

\begin{figure}[tb]
 \centering
 \includegraphics[width=\columnwidth]{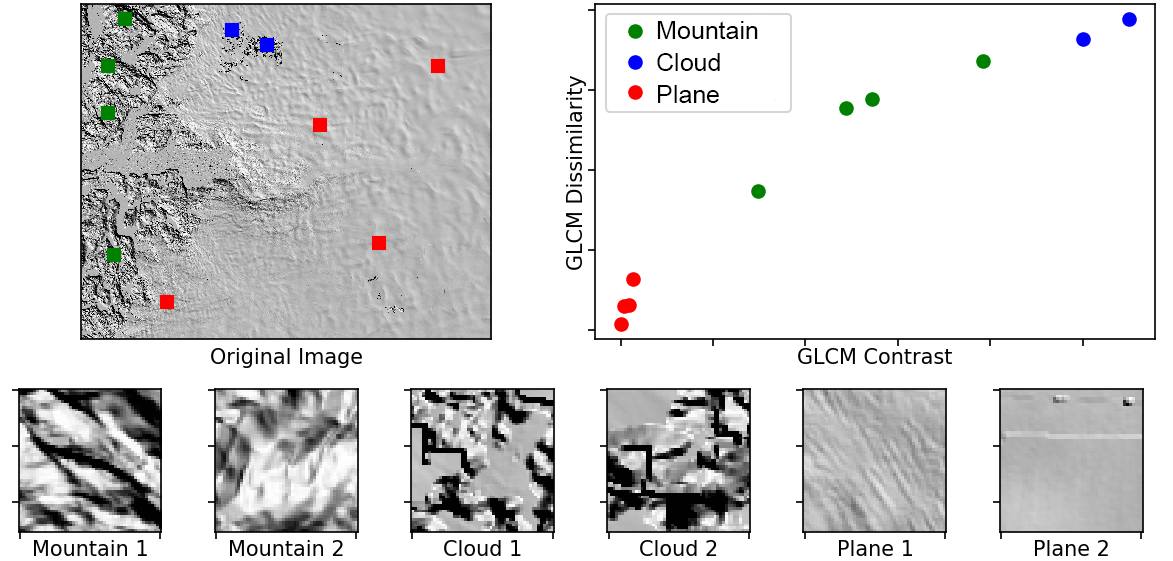}
 \caption{GLCM features for three main types of land covers.}
 \label{fig:glcm}
\end{figure}

Our initial machine learning model used these images as training data, but produced poor results where many discontinuous, individual pixels were identified as clouds rather than broad, connected areas. To resolve this issue, an additional second order textural analysis pre-processing step was added to create derivative data that considers the spatial relationship among the image pixels. A Gray Level Co-occurrence Matrix (GLCM) \cite{glcm} is an image representation which keeps track of different combinations of pixel values (gray levels) as they occur in an image, identifying various image texture features such as contrast, dissimilarity, homogeneity, and entropy. Figure \ref{fig:glcm} shows three of these features over different types of land covers. Calculating the GLCM requires specifying two parameters - the window size to use around each pixel, and the relationship direction, which is the distance vector between the reference pixel and the neighborhood pixel (often taken as a single unit distance in each of the $4$ directions left, right, up, and down). In order to consider both small-scale and large-scale texture features, $3$-, $5$-, and $15$-pixel window sizes were used to create three derivative datasets, to be used in an ensemble method of cloud mask prediction. Each of these datasets consisted of $4399$ $52$-channel textural ``images''. After the GLCM calculations, the images were normalized to be between $0$-$1$, as a best practice for machine learning.

\subsection{Deep Learning for Cloud Prediction}

\begin{figure}[tb]
 \centering
 \includegraphics[width=\columnwidth]{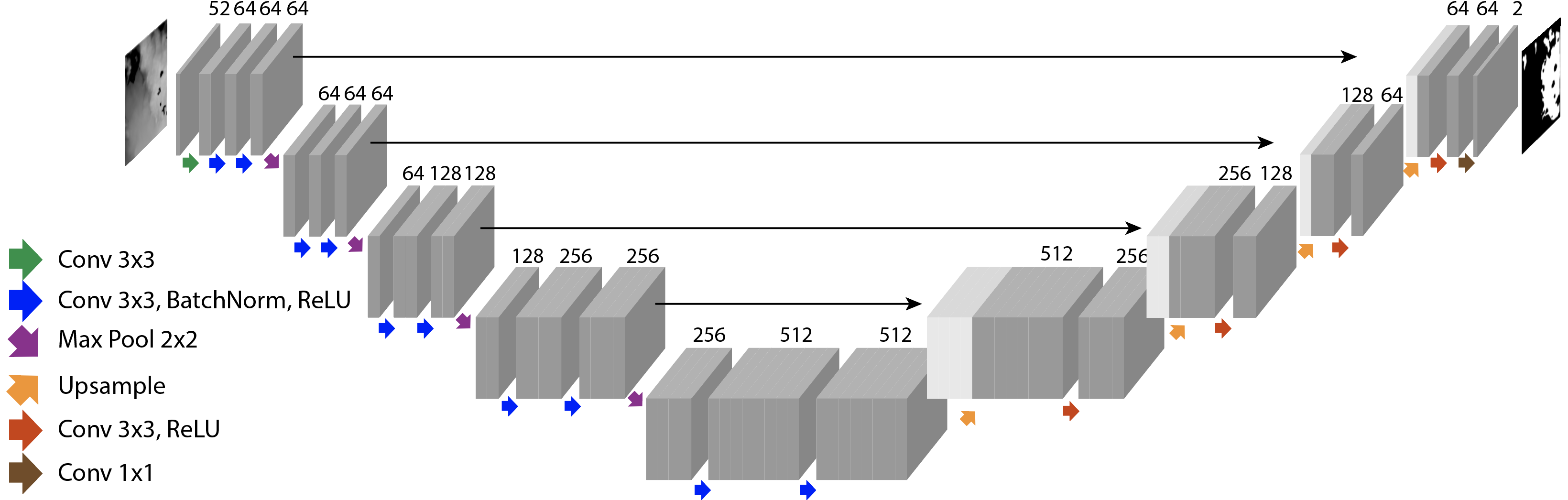}
 \caption{ The CloudFindr architecture, based on U-Net\cite{unet}. }
 \label{fig:unet}
\end{figure}
U-Net was selected as the basis for CloudFindr. Other architectures were considered - notably RS-Net \cite{JEPPESEN2019247} and MC-Net \cite{mcnet2021} - which are specialized use cases of the more basic underlying U-Net algorithm and are optimized for different use cases: RS-Net for spectral and MC-Net for multi-channel satellite imagery. U-Net was chosen as it is more generalized and allows for customization at a lower level. The CloudFindr architecture is outlined in Figure \ref{fig:unet}. The downstream branch consists of four convolutional blocks, each being a combination of two convolution and ReLU operations, followed by a maxpool to reduce the dimensions of the image by a factor of two (with stride $2$ and kernel size $2$). At the end of the downstream branch, the input is reduced to a size of width/16 by height/$16$ by $512$ features. The upstream branch consists of four upsampling convolutional blocks. Each block first upsamples the input by a factor of two using up-convolution followed by a ReLU operation, increasing the size of the input again by a factor of $16$. A final convolutional layer is applied to convert the resulting $16$ channels into $2$, followed by a softmax to obtain a probability for each class, ``cloud'' versus ``non-cloud''. The resulting image contains a pixel-wise confidence between $0$-$1$ for whether that pixel contains a cloud or not. This image is thresholded to produce discrete $0$ or $1$ values in the final output mask to give a prediction of ``cloud'' or ``no cloud''.

The dataset has a $60$-$20$-$20$ split between training-validation-testing. The hyperparameters of loss function, optimizer, learning rate, regulation, and number of epochs were tuned via control experiments. A combined evaluation of IoUs and segmentation results was performed after each experiment to determine if current variable value would be retained for next experiments. The optimal combination of parameters is found as: loss function weights = [$0.3$,$0.7$] to account for the imbalance between number of instances for each class, Adam optimizer with learning rate of $0.005$, no dropout regulation, and $200$ epochs. Both Adam and SGD optimizers were tested with learning rates between $0.005$ and $0.001$. The best results came from the use of Adam with a learning rate of $0.005$. 

\begin{figure}[tb]
 \centering
 \includegraphics[width=\columnwidth]{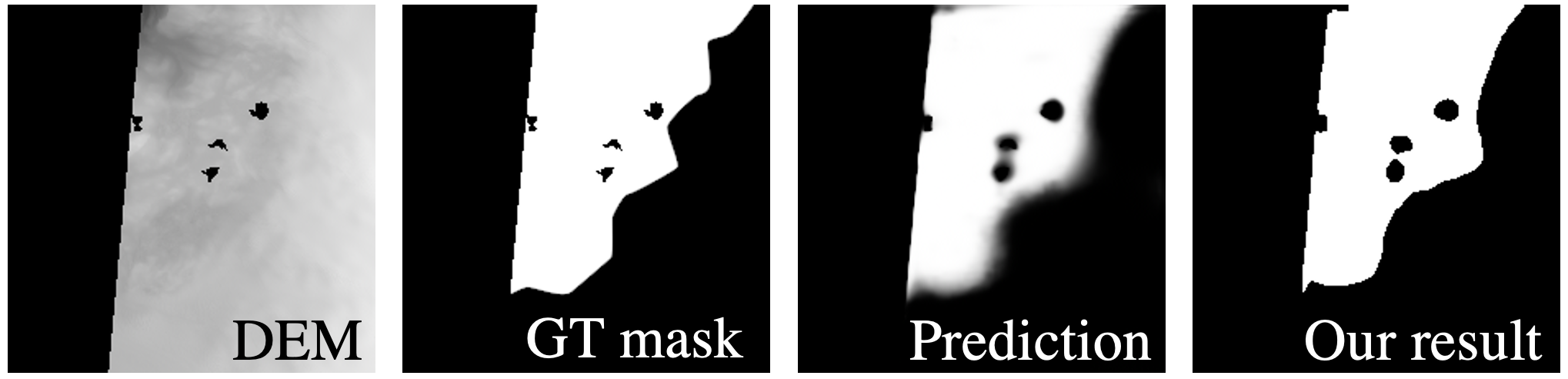}
 \caption{One example patch where it would be difficult for a casual observer to identify clouds, but the expert and machine learning prediction have closely-aligned results. From left to right: Input DEM patch, ground truth mask hand-drawn by an expert, confidence of prediction after ensemble voting, final thresholded predicted mask.}
 \label{fig:mask_prediction}
\end{figure}

Initially, the model was run on derivative datasets with GLCM window sizes of $3$, $5$, and $15$ with the aim of finding a single optimal window size. As designed, all resulting predictions skewed toward higher recall rather than higher precision and tended to over-label areas as ``clouds'' rather than under-labelling them. However by visually analyzing the output masks, it became clear that the three methods tended to agree with one another about the areas \textit{correctly} identified as clouds, but disagreed about the areas labelled \textit{incorrectly}. This inspired the use of an ensemble method for gathering the final result. The final prediction combines results from all three runs by multiplying the outputs together. The effect of this is that the overall confidence value is significantly reduced, but if any one of the runs predicts a $0$ value (predicting that there are no clouds present), this overrides any other predictions and a 0 value is placed in the final output mask. The multiplied confidence is thresholded with a value of $0.1$ to create the final binary cloud/non-cloud prediction. Figure \ref{fig:mask_prediction} shows one example patch prediction. 

When a cloud is mislabelled as a non-cloud, this most often appears around the perimeter of a correctly-labelled cloudy area. To account for this, a final post-processing step is applied to dilate the image masks with a kernel of size ($5$,$5$). This reduces the error around the edges of cloud regions, and creates masks that are slightly ``overdrawn'' similarly to how the human expert performed manual rotoscope labelling.

\section{Results}

\begin{figure}[tb]
 \centering
 \includegraphics[width=\columnwidth]{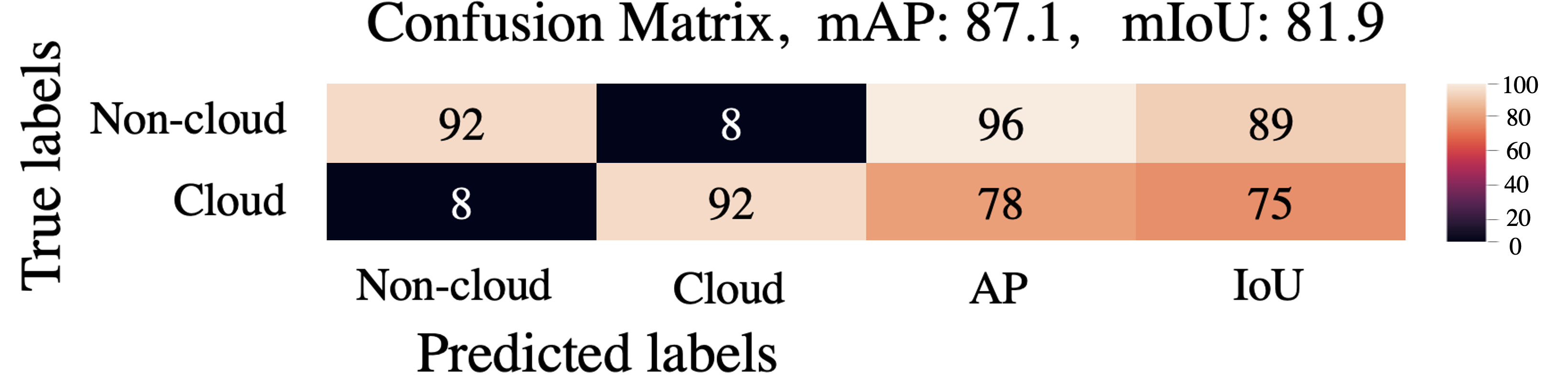}
 \caption{Confusion matrix showing the success of the predictions after all processing.}
 \label{fig:matrix}
\end{figure}

\begin{figure}[tb]
 \centering
 \includegraphics[width=\columnwidth]{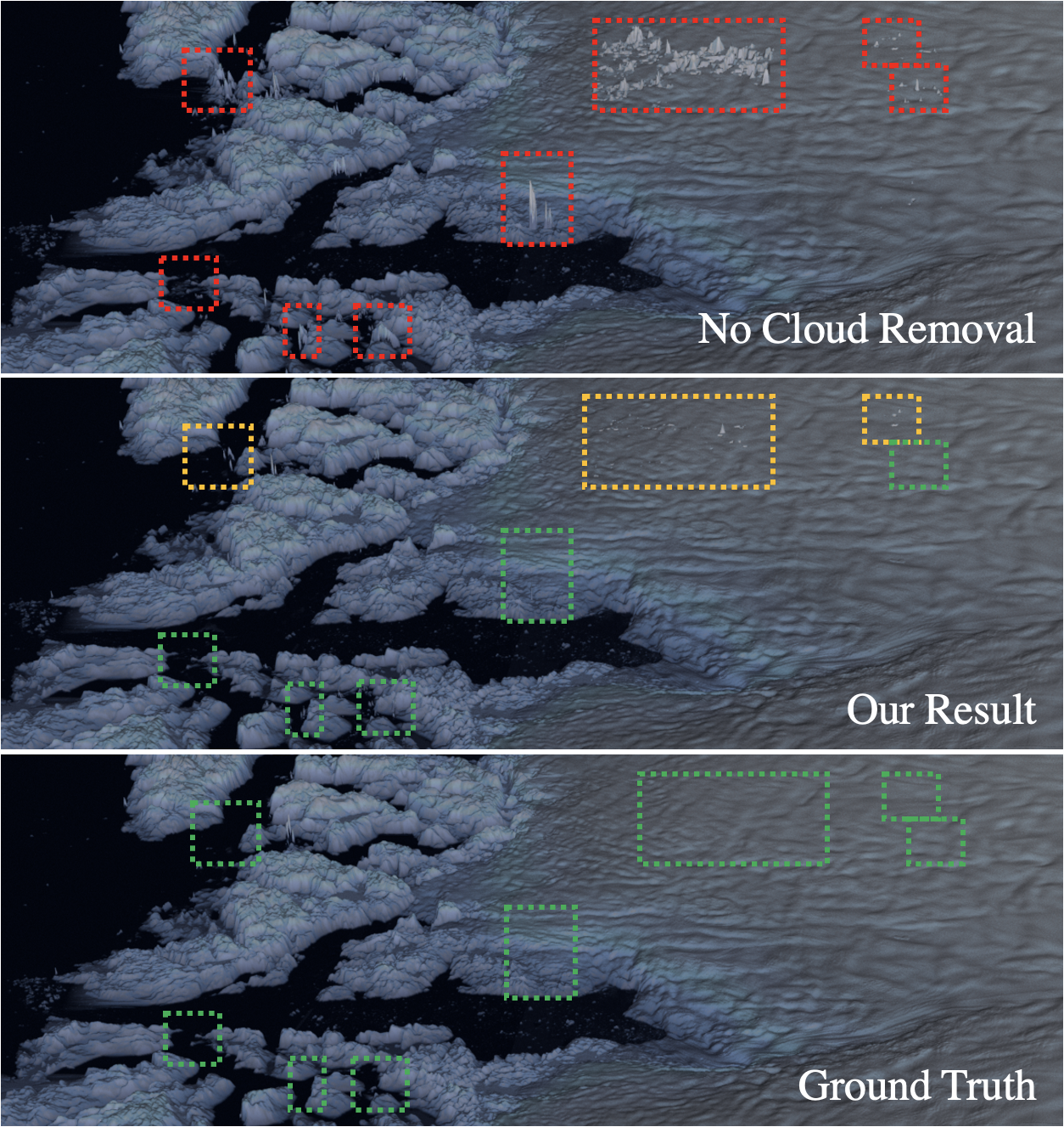}
 \caption{Images showing the same single frame of a final 3D render. Top: using no cloud mask. Middle: using cloud mask created via the method described here. Bottom: using masks created manually by a rotoscoping expert. Red boxes draw attention to areas with especially visible clouds; yellow boxes show that the clouds have been mostly removed; green boxes show that they have been entirely removed.}
 \label{fig:render}
\end{figure}

The neural network was trained on a GM200GL Quadro M6000 NVIDIA GPU for approximately $12$ hours. In the final result, the model was able to correctly identify cloudy DEM pixels $92$\% of the time. The mean average precision of the optimal model described above is $87.1$\% and the mean IoU is $81.9$\%, with a further breakdown for each class shown in Figure \ref{fig:matrix}.

The output of the described algorithm is $4399$ patches of size $224$x$224$ with values of 1 where there are likely clouds present, and 0 where there are not. These patches are stitched back together to create $978$ masks of size $3473$x$2840$ which can be multiplied against the $978$ DEMs of size $3473$x$2840$ around the Jakobshavn area. The DEM strips and masks are then accumulated to create the final DEMs to be used in the 3D cinematic rendering. Figure \ref{fig:render} shows how our result compares to the ground truth in final 3D rendered imagery, as well as what the render looks like without cloud removal. These renderings are created with the software Houdini\footnote{https://www.sidefx.com/products/houdini/}, where the DEM values are used to drive both the height and the color of the land. In this figure, the vast majority of the cloud artifacts have been removed, and the ones that have been missed are not as visually disturbing as the more prominent spikes.

\section{Conclusion and Future Work}
In this paper, we describe CloudFindr, a method of labelling pixels as ``cloud'' or ``non-cloud'' from a single-channel DEM image. We first extract textural features from the image with varying window sizes. We feed this derived data into a U-Net based model, trained on labelled data created by an expert, to create image segmentation predictions. The results have high accuracy as demonstrated both by metrics and by a 3D rendering created from the data.

In the future, we will plan a large hyperparameter tuning study including features at different sizes, learning rate, momentum, and batch size to optimize our results. Additionally, we would like to apply this method to other DEM datasets outside the Jakobshavn region of the ArcticDEM dataset, and also incorporate the time dimension into the training to differentiate between strips that are updating a previously-seen area from strips covering a new region.

\acknowledgments{
Thank you to Donna Cox, Bob Patterson, AJ Christensen, Saurabh Gupta, Sebastian Frith, and the reviewers. This work was supported by the Blue Waters Project, National Science Foundation, National Geospatial-Intelligence Agency, and Fiddler Endowment.}

\bibliography{cloudFindr}
\bibliographystyle{abbrv-doi}

\end{document}